\documentclass[letterpaper, 10 pt, onecolumn]{article}  

\usepackage{tabularx}
\usepackage{subfigure}
\usepackage{amsmath,amssymb}
\usepackage{latexsym}
\usepackage{psfrag}
\usepackage{epsfig}
\usepackage{graphicx}
\usepackage{pifont}
\usepackage{array}
\usepackage{collref}
\usepackage{enumerate}
\usepackage{cite}
\usepackage{amsmath}
\usepackage{amscd}
\usepackage{amssymb}
\usepackage{latexsym}
\usepackage{array}
\usepackage{tabularx}
\usepackage[ruled]{algorithm}
\usepackage{algorithmic}
\usepackage{epsfig,subfigure}
\usepackage{changepage}
\usepackage{stfloats}
\usepackage{multirow}
\usepackage{multirow}
\usepackage{bigstrut}
\usepackage{array}
\usepackage{graphicx}

\usepackage{algorithm}

\allowdisplaybreaks
\allowdisplaybreaks[1]

\begin{document}

\title{\bf A Brief Survey of Image Processing Algorithms in Electrical Capacitance Tomography}

\author{ Kezhi Li$^{1,2}$
\thanks{$^{1}$ Medical Research Council,
     Imperial College London, UK
        {\tt\small kezhili@imperial.ac.uk}}%
\thanks{$^{2}$ Magnetic Resonance Research Centre (MRRC),    University of Cambridge, UK}
}
%
     

\maketitle

\begin{abstract}
To study the fundamental physics of complex multiphase flow systems using advanced measurement techniques, especially the electrical capacitance tomography (ECT) approach, this article carries out an initial literature review of the ECT method from a point of view of signal processing and algorithm design. After introducing the physical laws governing the ECT system, we will focus on various reconstruction techniques that are capable to recover the image of the internal characteristics of a specified region based on the measuring capacitances of multi-electrode sensors surrounding the region. Each technique has its own advantages and limitations, and many algorithms have been examined by simulations or experiments. Future researches in 3D reconstruction and other potential improvements of the system are discussed in the end.

\end{abstract}

\section{Introduction}
The electrical capacitance tomography (ECT) technique belongs to the big family of tomography approaches. Similar ideas include computerised tomography (CT) and Electrical impedance tomography (EIT), etc., which have been familiar to people. Literally, the word tomography is derived from ``tomo-'' meaning ``to slice'' and ``graph'' having a meaning of ``image''. So it means to see the object we are interested in by having its images slice by slice.

Technically, tomography refers to the process of exploring the internal structure of an object through integral measurements without the need to invade the object. It is often perceived as an imaging tool for industrial monitoring or medical examination purposes. An applicable tomography technique often has two features: 1) non-invasive, which means no direct contact between the sensor and the region of interest; 2) non-intrusive, which means the measuring process does not change the nature of the object being examined.

ECT is a tomography technique that has these two features. It exploits measurements of the electrical capacitance obtained from multi-electrode sensors located surrounding the object or region of interest. This object or region usually refers to an industrial vessel or pipe containing two materials of different permittivities, and the measurement signals are used to reconstruct the permittivity distribution as well as the material distribution over the cross section by utilizing a suitable reconstruction algorithm \cite{Dyakowski-Applications}. So in practice ECT is often used to image cross-sections of the industrial process containing dielectric materials.

\begin{figure*}[t!]
   \centering
   \includegraphics[width=11.5cm]{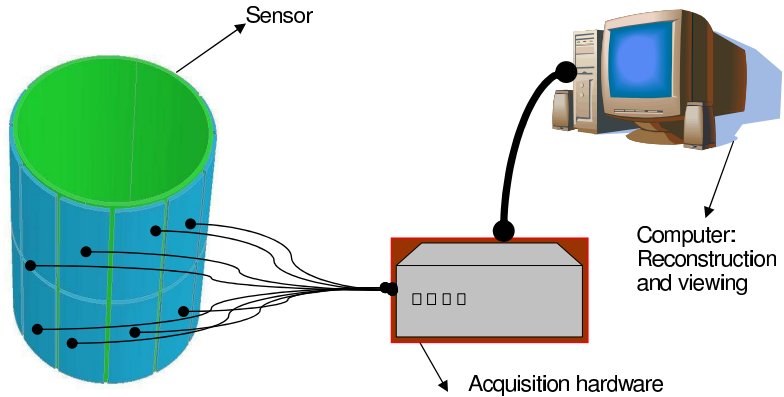}
      \caption{{An ECT system \cite{Marashdeh-thesis}.}}\label{fig:ECTSys}
    \end{figure*}

The technique of ECT has been developed for decades, and still attracts lots of attentions from both scientific researchers and experts from industry. The great interests in ECT for applications have been motivated by its high speed, high safety and low cost. It also suits for vessels of different sizes. A typical ECT system is shown in Fig.\ref{fig:ECTSys}, which consists of three main parts: a multi-electrode sensor, an acquisition hardware and a computer for hardware control and image processing. Apparently, it is non-invasive and non-intrusive. Specifically, the multi-electrode hardware in ECT typically has $n$ electrodes surrounding the wall of the process vessel as illustrated, and the number of independent capacitance measurement in such a configuration is $1/2\cdot n(n-1)$ due to the independent number of sensor pairs within $n$ sensors. The final objective is to recover the cross section or even 3D images by using these measurements. This problem is underdetermined, since number of measurements is far fewer than the number of pixels in the reconstructed image, and the highest recorded resolution normally does not exceed $3 \%$ of the imaging domain \cite{Marashdeh-thesis}. Thus various reconstruction algorithms are developed to cope with this difficulty.

In the remaining parts of the article, the physical laws of ECT system will be explained in Section 2; then we will put the main emphasise on the review of reconstruction algorithms in Section 3; finally Section 4 concludes and article and some future work will be discussed.

\section{Fundamentals of Electrical Capacitance Tomography}
As mentioned ECT provides a unique way of non-intrusive probes, and it allows for a detailed analysis of multiphase flow morphology. Here, the permittivity distribution inside a pipe, corresponding to the material distribution, is obtained from capacitance between all pairs of sensors located around the pipe's periphery. The relationship between the spatial distribution of the permittivity and the measured capacitances can be derived from Maxwell's equations and Gauss' law as
\begin{equation}
\nabla \cdot \mathbf{D}(\mathbf{r}) = \rho_v(\mathbf{r}),
\end{equation}
where $\mathbf{D}(\mathbf{r})$ is the electric flux density, $\rho_v(r)$ represents the volume charge density, and $\nabla \cdot$ is the divergence operator. Denoted the spatial permittivity distribution $\epsilon(\mathbf{r})$, the electric field intensity $\mathbf{E}(\mathbf{r})$ and the electric potential distribution $\phi(\mathbf{r})$, because $\mathbf{D}=\epsilon(\mathbf{r})\mathbf{E}(\mathbf{r})$ and $\mathbf{E}(\mathbf{r})=-\nabla_{\phi}(\mathbf{r})$ where $\nabla$ is the gradient operator, then we have
\begin{equation}
\mathbf{D}=-\epsilon(\mathbf{r})\nabla_{\phi}(\mathbf{r}).
\end{equation}
Since the total electric flux over all the electrodes surfaces is equal to zero, we have the Poisson's equation:
\begin{equation}
\nabla \cdot \left[ \epsilon(\mathbf{r})\nabla_{\phi}(\mathbf{r})\right] = 0,
\end{equation}
and the boundary conditions are $\phi=V_c$ for excited electrode and $\phi=0$ for other electrodes. For the two-dimensional case $\mathbf{r}=(x,y)$, the relationship between the capacitance and permittivity distribution can be expressed by the following equation:
\begin{equation}\label{eq:C}
C=\frac{Q}{V_c} = -\frac{1}{V_c} \oint_S \epsilon(x,y) \nabla_{\phi}(x,y)\text{d}s,
\end{equation}
where $Q$ is the total charge, $S$ denotes the closed line of the electrical field, $\epsilon(x,y)$ is the permittivity distribution in the sensing field, and $V_c$ is the potential difference between two electrodes forming the capacitance.

\begin{figure*}[t!]
   \centering
   \includegraphics[width=12cm]{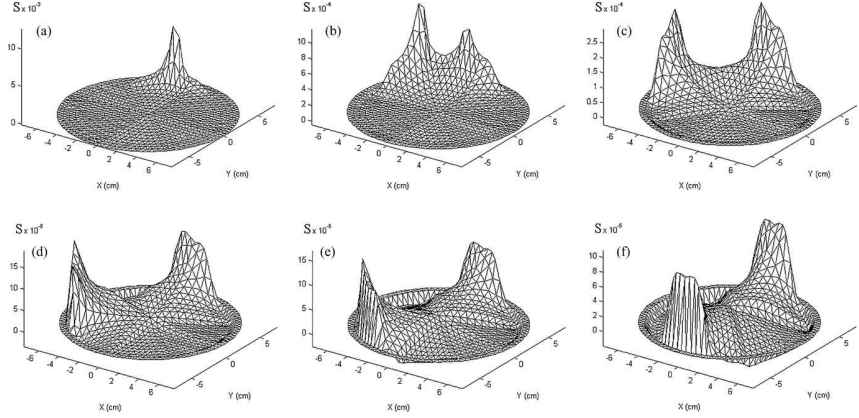}
      \caption{Sensitivity maps for the conventional sensor \cite{Olmos-Development}.}\label{fig:Sensitivity}
    \end{figure*}

Equation (\ref{eq:C}) can be simplified in some circumstances. If an ideal parallel-plate sensor with homogeneous permittivity distribution, it becomes
\begin{equation}
C= \epsilon_0 \epsilon_r \frac{A}{d},
\end{equation}
where $\epsilon_0$ is the permittivity of vacuum, $\epsilon_r$ is the relative permittivity of the material inside the sensor, $A$ represents the area of the plates and $d$ is the distance between the two plates. However, the geometry distribution of ECT sensors is more complicated. In (\ref{eq:C}) ${\phi}(x,y)$ is also a function of $\epsilon$. Therefore the capacitance between electrode combinations can be considered as a function of permittivity distribution $\epsilon$:
\begin{equation}
C= f(\epsilon),
\end{equation}
where $f$ is a non-linear function, and elements of $C$ are not redundant elements of electrode pairs $\left[ C_{1,2}, C_{1,3}, \cdots, C_{1,n}, C_{2,3}\cdots C_{N-1,N} \right]$. If we take the differential on both sizes, the change will be
\begin{equation}\label{eq: triangleC}
\triangle C = \frac{\text{d} f}{\text{d} \epsilon}(\triangle \epsilon) + O ((\triangle \epsilon)^2).
\end{equation}
due to the Taylor expansion, where $\frac{\text{d} f}{\text{d} \epsilon}$ is the sensitivity of the capacitance versus permittivity distribution, and $O ((\triangle \epsilon)^2)$ represents the higher order terms of $(\triangle \epsilon)^2$. Because $\triangle \epsilon$ is usually small, the high order term is often neglected. Then (\ref{eq: triangleC}) can be linearized in a matrix form:
\begin{equation}\label{eq:linearC}
\triangle \mathbf{C} = \mathbf{J} \triangle \mathbf{\epsilon},
\end{equation}
where $\triangle \mathbf{C} \in \mathbb{R}^M$, $\mathbf{J}\in \mathbb{R}^{M \times N}$ is a Jacobian/sensitivity matrix denoting the sensitivity distribution for each electrode pair, and $\triangle \mathbf{\epsilon} \in \mathbb{R}^{N}, N \gg M$. As a result, the non-linear forward problem has been formulated to a linear approximation. Generally in ECT, (\ref{eq:linearC}) can be written in a normalized form
\begin{equation}\label{eq:lambda}
\mathbf{\lambda} = \mathbf{S} \mathbf{g},
\end{equation}
where $\mathbf{\lambda} \in \mathbb{R}^M$ is the normalized capacitance vector, $\mathbf{S}\in \mathbb{R}^{M \times N}$ is the Jacobian matrix of the normalized capacitance with respect to the normalized permittivities, which gives a sensitivity map for each electrode pair, and $\mathbf{g} \in \mathbb{R}^{N}, N \gg M$ is the normalized permittivity vector, which can be visualized by the colour density of the image pixels. The sensitivity maps for the conventional sensor are shown in Fig. \ref{fig:Sensitivity}. Because there are $n$ electrode pairs, $M$ should be $1/2\cdot n(n-1)$. The objective of the reconstruction algorithm of ECT is to recover $\epsilon(x,y)$ from measured capacitance vector $C$. While in the discrete linear model, it is to estimate $\mathbf{g}$ given $\mathbf{\lambda}$, and $\mathbf{S}$ is seen as a constant matrix determined in advance for simplicity.

There are several difficulties of the reconstruction problem. Firstly, (\ref{eq:lambda}) is under-determined so the solution is not unique \cite{Marashdeh-thesis}. Secondly, (\ref{eq:C}) is even harder to solve since it is ill posed and not linear, its linear form (\ref{eq:lambda}) is very sensitive to disturbances of $\lambda$. At last, $\mathbf{S}$ is not always constant in practice. It varies for different permittivity distributions. In the following section we will introduce several reconstruction algorithms that can overcome these obstacles to some extent.

\section{Reconstruction Algorithms}

Generally speaking the reconstruction algorithms can be categorized in two groups: direct algorithms and iterative algorithms. Direct algorithms are faster and hold the assumption that the sensitivity does not change as a function of permittivity distribution. Iterative algorithms are slower with regard to the speed but usually have better performances. Firstly we will introduce several direct algorithms, including linear back projection (LBP), algorithm based on singular value decomposition (SVD) and algorithm based on Tikhonov regularization.

\subsection{ Direct Algorithms}
\subsubsection{Linear Back Projection}

Linear back projection (LBP) technique is based mainly on the sensitivity matrix model and assume that sensitivity matrix is invariant. Each element of the matrix $\mathbf{S}$ is obtained from the response of a pair of electrodes to a perturbation of high electrical permittivity in the imaging domain. Then the elements are normalized by the relative capacitance:
\begin{equation}
\mathbf{\lambda}_{i,j} = \frac{C_{i,j}- C_{i,j}^l}{C_{i,j}^h- C_{i,j}^l},
\end{equation}
where $\mathbf{\lambda}_{i,j}$ is the element of $\lambda$ with the $j$th capacitance pair and $i$th pixel, $C_{i,j}$ is the measured capacitance, and $C_{i,j}^h,C_{i,j}^l$ are the capacitances when the pipe is filled with high or low permittivity materials, respectively. $\mathbf{\lambda}$ is the vecterized $\mathbf{\lambda}_{i,j}$ for all $i,j \in 1,\cdots, n, i\neq j$ The forward problem is formulated in (\ref{eq:lambda}). If the inverse of $\mathbf{S}$ exists, the inverse problem has a solution:
\begin{equation}
\mathbf{g} = \mathbf{S}^{-1} \mathbf{\lambda}.
\end{equation}
When the inverse of $\mathbf{S}$ does not exist, the image vector can be obtained from a linear mapping from the capacitance vector using the transpose of sensitivity matrix as
\begin{equation}
\mathbf{g} = \mathbf{S}^T \mathbf{\lambda},
\end{equation}
or a normalized version:
\begin{equation}
\mathbf{g} = \frac{\mathbf{S}^T \mathbf{\lambda}}{\mathbf{S}^T \mathbf{\mathbf{u}_{\lambda}}}, \mathbf{u}_{\lambda} = \left[1,1,\cdots, 1 \right],
\end{equation}
where $\mathbf{u}_{\lambda}$ is an identity vector.

\subsubsection{Approach based on SVD}
If we consider the errors occurred in the measurement process, (\ref{eq:lambda}) becomes
\begin{equation}\label{eq:lambdaError}
\mathbf{\lambda} = \mathbf{S} \mathbf{g}+\mathbf{e},
\end{equation}
where $\mathbf{e}$ is the capacitance measurement error which satisfies certain distributions, such as Gaussian random distribution. Then the least square solution to (\ref{eq:lambdaError}) would be
\begin{equation}\label{eq:g_inv}
\mathbf{g} = (\mathbf{S}^T\mathbf{S})^{-1} \mathbf{S}^T \mathbf{\lambda}.
\end{equation}
However when the $(\mathbf{S}^T\mathbf{S})^{-1}$ is not computable, a pseudoinverse operator can be formulated as follows:
\begin{equation}
\mathbf{S} = \mathbf{U \Sigma V}^T,
\end{equation}
which is the SVD of $\mathbf{S}$. Then the pseudoinverse of matrix $\mathbf{S}$ can be calculated as
\begin{equation}\label{eq:S1}
\mathbf{S}^{\intercal}=\mathbf{V} \mathbf{\Sigma}^{-1} \mathbf{U}^T,
\end{equation}
where $\mathbf{\Sigma}^{-1}$ is given by
\begin{equation}
\mathbf{\Sigma}^{-1} = \text{diag}\left[1/\sigma_1, 1/\sigma_2, \cdots, \right],
\end{equation}
where $\sigma_i, i \in \{1,2,\cdots \}$ are the singular values of $\mathbf{S}$. Then the reconstruction equation becomes:
\begin{equation}
\mathbf{g} = \mathbf{S}^{\intercal} \mathbf{\lambda}.
\end{equation}

We can also implement a truncated SVD to improve the performances of situations that an obvious gap exists between singular values. The truncated version of SVD also utilize (\ref{eq:S1}) to calculate $\mathbf{S}^{\intercal}$ but with a modified version of $\mathbf{\Sigma}^{-1}$:
\begin{equation}
\mathbf{\Sigma}^{-1} = \text{diag}\left[w_1/\sigma_1, w_2/\sigma_2, \cdots, \right]
\end{equation}
where $w_i=  \frac{\sigma_i^2}{\sigma_i^2+\mu},i \in \{1,2, \cdots\}$, $\mu$ is a positive regularization parameter. The algorithm based on truncated SVD is less sensitive to high frequency noise involved in the measurements.

\subsubsection{Tikhonov regularization}

Regularization methods have been developed for many years to solve ill-posed inverse problems. Specifically, Tikhonov regularization is verified as an efficient method to solve the ill-posed problems in ECT. To cope with the situations that $\mathbf{S}^T\mathbf{S}$ is not invertible, the equation (\ref{eq:g_inv}) can be modified based on the standard Tikhonov regularization procedure as
\begin{equation}\label{eq:g_inv2}
\mathbf{g} = (\mathbf{S}^T\mathbf{S}+\mu \mathbf{I})^{-1} \mathbf{S}^T \mathbf{\lambda},
\end{equation}
where $\mathbf{I}$ is an identity matrix and $\mu$ is a positive regularization parameter. In practice, the quality of reconstructed images depends strongly on $\mu$. In mathematics, (\ref{eq:g_inv2}) can be derived from a more general form of Tikhonov regularization problem which tries to optimize:
\begin{equation}
\arg \min_{\mathbf{g}} ||\mathbf{Sg}-\mathbf{\lambda}||_2^2 + \mu ||\mathbf{g-\hat{g}}||^2_2
\end{equation}
where $\hat{\mathbf{g}}$ is the estimated $\mathbf{g}$. Because $\hat{\mathbf{g}}$ is often difficult to obtain, setting $\hat{\mathbf{g}}$ as zero gives the standard Tickhonov regularization in (\ref{eq:g_inv2}).

\subsection{Iterative algorithms}

\subsubsection{Iterative Back Projection Algorithm}
Linear back projection algorithm is based on the conventional linear back projection process to obtain the initial image. Then the values of capacitance is updated from a forward projection. After that, the difference between the measured and estimated capacitances is calculated and back projected. These steps are iterated until the difference approaches zero. In mathematics, the algorithm can be stated as:
\begin{equation}
\begin{split}
\mathbf{g}^{0} &= \mathbf{S}^T \mathbf{\lambda}  \\
\mathbf{g}^{k+1}&=\mathbf{g}^{k} + \alpha \mathbf{S}^T(\mathbf{\lambda}-\mathbf{S}\mathbf{g}^{k})
\end{split}
\end{equation}
where $k=0,1,\cdots$ denotes the number of iterations, $\alpha$ is a relaxation parameter which
\begin{equation}
\alpha^{k} := \frac{||\mathbf{S}^T (\mathbf{\lambda}-\mathbf{S}\mathbf{g}^{k})||^2_2}{||\mathbf{SS}^T (\mathbf{\lambda}-\mathbf{S}\mathbf{g}^{k})||^2_2}.
\end{equation}
Here, the sensitivity matrix remains constant for all iterations. The sensitivity matrix also can be updated during the iterations as an alternative approach.
\subsubsection{Landwater Iteration and Steepest Descent Method}
Landwater iteration is widely used in the optimization theory that leverages the steepest gradient descent method to minimize the cost function. In our case the cost function is to minimize $\frac{1}{2} ||\mathbf{Sg}- \mathbf{\lambda}||^2_2$, e.g. to minimize
\begin{equation}
\begin{split}
f(\mathbf{g}) &= \frac{1}{2} (\mathbf{Sg}-\mathbf{\lambda})^T(\mathbf{Sg}-\mathbf{\lambda}) \\
&=\frac{1}{2}(\mathbf{g}^T\mathbf{S}^T\mathbf{Sg}-2\mathbf{g}^T\mathbf{S}^T\mathbf{\lambda}+\mathbf{\lambda}^T\mathbf{\lambda}). \end{split}
\end{equation}
The gradient of $f(\mathbf{g})$ is
\begin{equation}
\nabla f(\mathbf{g})=\mathbf{S}^T\mathbf{S}g -\mathbf{S}^T\mathbf{\lambda}= \mathbf{S}^T (\mathbf{Sg}-\mathbf{\lambda}),
\end{equation}
which means we choose the direction that $f(\mathbf{g})$ decreases most quickly due to the gradient descent method used here. Therefore the new image will be
\begin{equation}
\mathbf{g}^{k+1}= \mathbf{g}^k - \alpha^k \nabla f(\mathbf{g}^k)=  \mathbf{g}^k - \alpha^k\mathbf{S}^T (\mathbf{Sg}^{k}-\mathbf{\lambda}),
\end{equation}
where $\alpha_k$ is a positive value determining the step size.

\subsubsection{Iterative Soft Thresholding of Total Variation}
In contrast to the previous methods, besides the least squares the this approach uses the total variation (TV) term in the cost function \cite{Chandrasekera-TotalVar}. Then the reconstruction problem in terms of the image gradient is thus:
\begin{equation}
\min_{\mathbf{g}}||\mathbf{\lambda}-\mathbf{Sg}||^2_2+\alpha||\mathbf{g}||_{TV}
\end{equation}
where $\mathbf{g}=\{\mathbf{g}_x, \mathbf{g}_x\}$ represents the pixels along $x,y$ directions, respectively. The $l_{TV}$ norm here denotes the total variation of the pixels along $x$ and $y$ direction. Because the object in the pipe is normally not with complicate shape, it is reasonable to assume that the total variation of $\mathbf{g}$ is small. Combining the steepest descent idea from the previous algorithm, the iteration of soft thresholding (IST) is similar to the Landweber algorithm with the form:
\begin{equation}
\begin{split}
\mathbf{g}_x^{k+1}&= \mathbf{g}_x^{k} + \beta \mathbf{d}_x,\\
\mathbf{g}_y^{k+1}&= \mathbf{g}_y^{k} + \beta \mathbf{d}_y,
\end{split}
\end{equation}
where $\mathbf{d}_x,\mathbf{d}_y$ are the descent directions with step length $\beta$. Then 2D soft threshold update is applied to each element of $\mathbf{g}_x$ and $\mathbf{g}_y$. The algorithm proceeds until a convergence is arrived.

\subsubsection{Other Algorithms}
Except for the algorithms introduced above, there are lots of other iterative algorithms, such as iterative Tikhonov methods (ITM) \cite{Peng-UsingReg}, algebraic reconstruction technique (ART), algorithm based on neural network, accelerated model-based iteration \cite{Michailovich-AnIter, Wang-ANewAlter,Beck-FastGraBasedAlg}, nonquadratic regularizers algorithms include wavelet representations \cite{Guerquin-AFastWavelet}, sparse regression \cite{Donoho-OptimallySparse} and total variation \cite{Rudin-NonlinearTV, Soleimani-NonlinearImage}, as well as algorithms making use of the novel compressive sensing (CS) theory \cite{donoho-cs} etc. The details are not expanded and people have interests may refer to \cite{Yang-ImageRec, Beck-FastGraBasedAlg,Chandrasekera-TotalVar}.

\section{Conclusion and Future Work}
The ECT is still an attractive research field and people have done impressive work in the area of hardware implementation and reconstruction algorithm design. Recently compressive sensing (CS) is also considered as an efficient approach in developing algorithms, because the sparsity and the structure information can be effectively exploited in both theory and practice by using various reconstruction algorithms. Meanwhile 3-dimensional ECT has gained interests due to its potential to generate volumetric images. Our future work will focus on these two problems and try to address the issues by leveraging advanced signal processing and visualization techniques.

\section*{Acknowledgement}
This work was partially supported by the European Research Council
Grant Reference: EP/K008218/1.

\bibliographystyle{IEEEtran}

\begin{thebibliography}{10}
\providecommand{\url}[1]{#1}
\csname url@samestyle\endcsname
\providecommand{\newblock}{\relax}
\providecommand{\bibinfo}[2]{#2}
\providecommand{\BIBentrySTDinterwordspacing}{\spaceskip=0pt\relax}
\providecommand{\BIBentryALTinterwordstretchfactor}{4}
\providecommand{\BIBentryALTinterwordspacing}{\spaceskip=\fontdimen2\font plus
\BIBentryALTinterwordstretchfactor\fontdimen3\font minus
  \fontdimen4\font\relax}
\providecommand{\BIBforeignlanguage}[2]{{%
\expandafter\ifx\csname l@#1\endcsname\relax
\typeout{** WARNING: IEEEtran.bst: No hyphenation pattern has been}%
\typeout{** loaded for the language `#1'. Using the pattern for}%
\typeout{** the default language instead.}%
\else
\language=\csname l@#1\endcsname
\fi
#2}}
\providecommand{\BIBdecl}{\relax}
\BIBdecl

\bibitem{Dyakowski-Applications}
T.~Dyakowski, L.~F. Jeanmeure, and A.~J. Jaworski, ``Applications of electrical
  tomography for gas–solids and liquid–solids flows — a review,''
  \emph{Powder Technology}, vol. 112, no.~3, pp. 174 -- 192, 2000.

\bibitem{Marashdeh-thesis}
Q.~Marashdeh, ``Advances in electrical capacitance tomography,'' Ph.D.
  dissertation, Ohio State University, 2006.

\bibitem{Olmos-Development}
A.~M. Olmos, M.~Carvajal, D.~Morales, A.~García, and A.~Palma, ``Development
  of an electrical capacitance tomography system using four rotating
  electrodes,'' \emph{Sensors and Actuators A: Physical}, vol. 148, no.~2, pp.
  366 -- 375, 2008.

\bibitem{Chandrasekera-TotalVar}
T.~Chandrasekera, Y.~Li, J.~Dennis, and D.~Holland, ``Total variation image
  reconstruction for electrical capacitance tomography,'' in \emph{IEEE
  International Conference on Imaging Systems and Techniques (IST),}, Jul.
  2012, pp. 584--589.

\bibitem{Peng-UsingReg}
L.~H. Peng, H.~Merkus, and B.~Scarlett, ``Using regularization methods for
  image reconstruction of electrical capacitance tomography,'' \emph{Part.
  Part. Syst. Charact.}, vol.~17, pp. 96--104, 2000.

\bibitem{Michailovich-AnIter}
O.~Michailovich, ``An iterative shrinkage approach to total-variation image
  restoration,'' \emph{IEEE Transactions on Image Processing}, vol.~20, no.~5,
  pp. 1281--1299, May 2011.

\bibitem{Wang-ANewAlter}
\BIBentryALTinterwordspacing
Y.~Wang, J.~Yang, W.~Yin, and Y.~Zhang, ``A new alternating minimization
  algorithm for total variation image reconstruction,'' \emph{SIAM Journal on
  Imaging Sciences}, vol.~1, no.~3, pp. 248--272, 2008. [Online]. Available:
  \url{http://dx.doi.org/10.1137/080724265}
\BIBentrySTDinterwordspacing

\bibitem{Beck-FastGraBasedAlg}
A.~Beck and M.~Teboulle, ``Fast gradient-based algorithms for constrained total
  variation image denoising and deblurring problems,'' \emph{IEEE Transactions
  on Image Processing}, vol.~18, no.~11, pp. 2419--2434, Nov. 2009.

\bibitem{Guerquin-AFastWavelet}
M.~Guerquin-Kern, M.~Haberlin, K.~Pruessmann, and M.~Unser, ``A fast
  wavelet-based reconstruction method for magnetic resonance imaging,''
  \emph{IEEE Transactions on Medical Imaging}, vol.~30, no.~9, pp. 1649--1660,
  Sept 2011.

\bibitem{Donoho-OptimallySparse}
D.~L. Donoho and M.~Elad, ``Optimally sparse representation in general
  (nonorthogonal) dictionaries via $l_1$ minimization,'' \emph{Proc. Nat. Acad.
  Sci.}, vol. 100, no.~5, pp. 2197--2202, Mar. 2003.

\bibitem{Rudin-NonlinearTV}
L.~I. Rudin, S.~Osher, and E.~Fatemi, ``Nonlinear total variation based noise
  removal algorithms,'' \emph{Physica D: Nonlinear Phenomena}, vol.~60, no.
  1–4, pp. 259 -- 268, 1992.

\bibitem{Soleimani-NonlinearImage}
M.~Soleimani and W.~R.~B. Lionheart, ``Nonlinear image reconstruction for
  electrical capacitance tomography experimental data using,''
  \emph{Measurement Science \& Technology}, vol.~16, no.~10, pp. 1987--1996,
  2005.

\bibitem{donoho-cs}
D.~L. Donoho, ``Compressed sensing,'' \emph{{IEEE} Trans. Inf. Theory},
  vol.~52, pp. 1289--1306, Jul. 2006.

\bibitem{Yang-ImageRec}
W.~Q. Yang and L.~Peng, ``Image reconstruction algorithms for electrical
  capacitance tomography,'' \emph{Measurement Science and Technology}, vol.~14,
  no.~1, pp. R1--R3, 2003.

\end{thebibliography}

\end{document}